%% file: main.tex
\documentclass[10pt,twocolumn,letterpaper]{article}

\usepackage{iccv}
\usepackage{times}
\usepackage{epsfig}
\usepackage{graphicx}
\usepackage{amsmath}
\usepackage{amssymb}

\usepackage{array}
\usepackage{booktabs}
\usepackage{xcolor}
\usepackage{cite}
\usepackage{multirow}
\usepackage{soul}
\usepackage{algorithm}
\usepackage{algpseudocode}

\newcommand{\xhdr}[1]{\vspace{5pt} \noindent {\textbf{#1}}}

\usepackage[pagebackref=true,breaklinks=true,letterpaper=true,colorlinks,bookmarks=false]{hyperref}

 \iccvfinalcopy 

\def\iccvPaperID{***} 

\ificcvfinal\fi
\begin{document}

\title{StartNet: Online Detection of Action Start in Untrimmed Videos}

\author{
    Mingfei Gao$^1$\thanks{Work done
    when the author was an intern at Salesforce Research.}
    \hspace{0.35cm} Mingze Xu$^2$
    \hspace{0.35cm} Larry S. Davis$^1$
    \hspace{0.25cm} Richard Socher$^3$
    \hspace{0.25cm} Caiming Xiong$^3$\thanks{Corresponding author.}\\[.6ex]
    $^1$University of Maryland
    \hspace{0.25cm} $^2$Indiana University
    \hspace{0.25cm} $^3$Salesforce Research \\
    {\tt\small  \{mgao,lsd\}@umiacs.umd.edu, mx6@indiana.edu, \{rsocher,cxiong\}@salesforce.com}
}

\maketitle

\begin{abstract}
    We propose StartNet to address Online Detection of Action Start
    (ODAS) where action starts and their associated categories are
    detected in untrimmed, streaming videos. Previous methods aim to
    localize action starts by learning feature representations that
    can directly separate the start point from its preceding background.
    It is challenging due to the subtle appearance difference near
    the action starts and the lack of training data. Instead, StartNet
    decomposes ODAS into two stages: action classification
    (using ClsNet) and start point localization (using LocNet).
    ClsNet focuses on per-frame labeling and predicts action score
    distributions online.
    Based on the predicted action scores of the past and current
    frames, LocNet conducts class-agnostic start detection
    by optimizing long-term localization rewards using
    policy gradient methods.
    The proposed framework is validated on two
    large-scale datasets, THUMOS'14 and ActivityNet.
    The experimental results show that StartNet significantly
    outperforms the state-of-the-art by $15\%$-$30\%$ p-mAP under
    the offset tolerance of $1$-$10$ seconds on THUMOS'14,
    and achieves comparable performance on ActivityNet
    with $\times 10$ smaller time offset.
\end{abstract}

\input{introduction}
\input{related_work}
\input{method}

\input{experiments}
\input{conclusion}

{\small
\bibliographystyle{ieee}
\bibliography{egbib}
}

\end{document}

%% file: introduction.tex
\section{Introduction}
\label{sec: intro}

\begin{figure}[t]
    \begin{center}
        \includegraphics[width=1.0\linewidth]{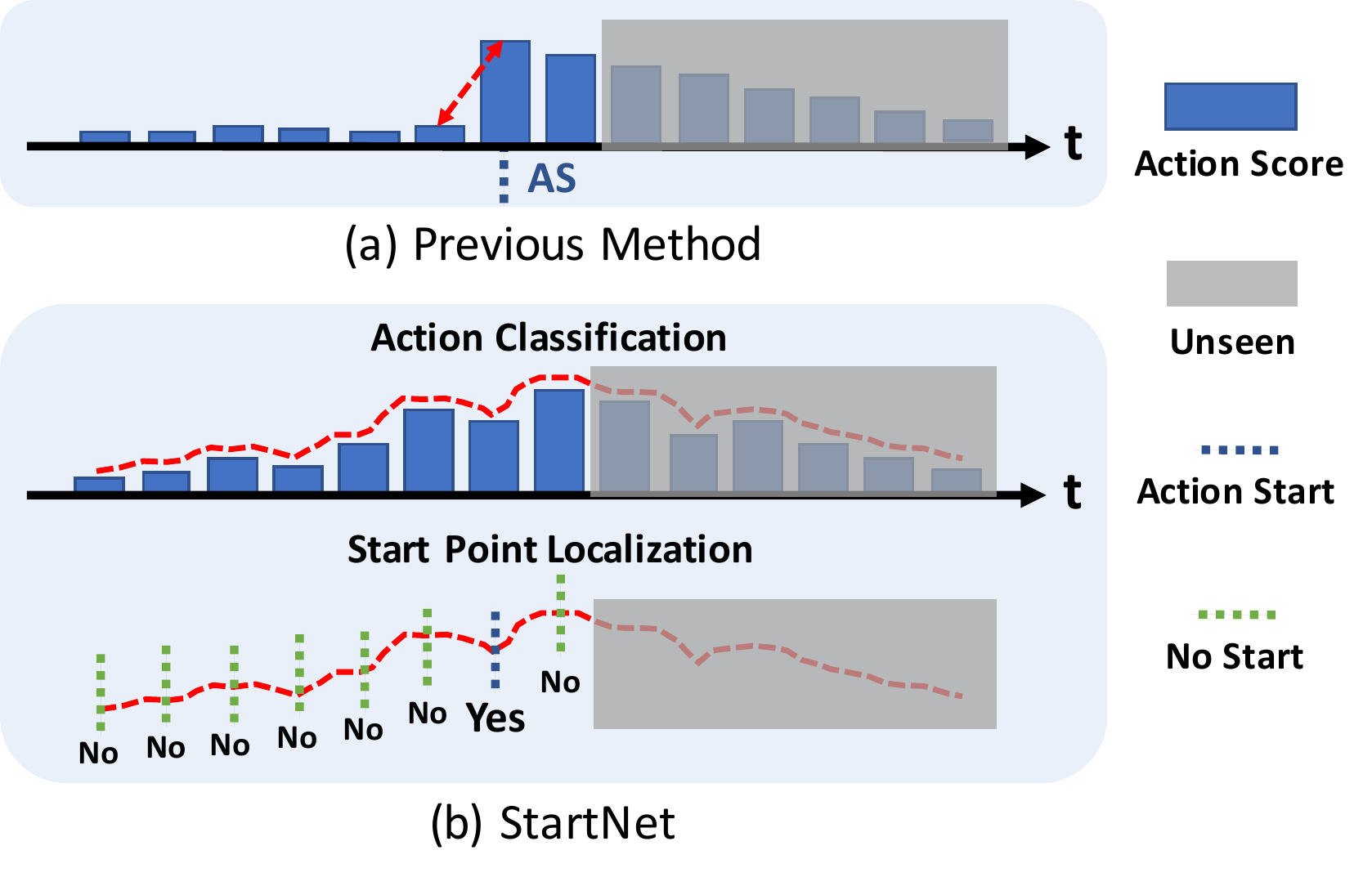}
    \end{center}
    \vspace{-10pt}
    \caption{
        Comparison between (a) the previous method~\cite{shou2018online}
        and (b) the proposed framework.~\cite{shou2018online} aims to
        generate an action score sequence which produces low
        score for background and high score for the correct action
        immediately when the action starts-- like a step function. We
        propose a two-stage framework: the first stage only focuses on
        per-frame action classification and the second stage learns to
        localize the start points given the historical trend of the
        action scores generated by the first stage.
    }
    \label{fig:idea}
       \vspace{-10pt}
\end{figure}

Temporal action localization (TAL) in untrimmed videos has been widely
studied in offline settings, where start and end times of an action are
recognized after the action is fully observed~\cite{shou2016temporal,zhao2017temporal,dai2017temporal,buch2017sst,gao2017turn,chao2018rethinking}.
With the emerging applications that require identifying actions in real time,
\eg, autonomous driving, surveillance system, and collaborative robots,
online action detection (OAD) methods~\cite{de2016online,gao2017red,shou2018online,xu2018temporal}
have been proposed. They typically pose the TAL problem as a per-frame class labeling task. 

However, in some time-sensitive scenarios, detecting accurate
action starts in a timely manner is more important than
successfully detecting every frame containing actions.
For example, an autonomous driving car needs to detect the start of
``pedestrian crossing" as soon as it happens to avoid collision;
a surveillance system should generate alert as soon as a dangerous event
is initiated. Online Detection of Action Start (ODAS) was proposed to
address this problem specifically~\cite{shou2018online}. Instead of
classifying every frame, ODAS detects the occurrence and category of
an action start as soon as possible. Thus, it addresses two sub-tasks:
(i) if an action starts at time $t$ and (ii) its associated action class.

The existing method~\cite{shou2018online} handles the two sub-tasks
jointly by training a classification network that is capable of localizing
the starts of different action classes. The network attempts to
make the representation of a start point close to that of its associated
action class and far from its preceding background. As shown in Fig.~\ref{fig:idea} (a), the network is encouraged to react
immediately when an action starts. However, it is hard to achieve this
goal due to the subtle appearance difference near start points and the lack of labeled training data (one action only contains one start point).

Our method is inspired by three key insights. First, decomposing a complex
task properly allows sub-modules to focus on their own sub-tasks and makes
the learning process easier. A good example is the success of the two-stage
object detection framework~\cite{girshick2014rich,girshick2015fast,ren2015faster}.
Second, as mentioned in~\cite{girshick2014rich}, when training data is scarce,
learning from a representation that is pre-trained on an auxiliary task
may lead to a significant performance boost. Third, OAD (per-frame labeling)
is very related to ODAS. Compared to the scarce labeled data of action starts,
the amount of per-frame action labels is much larger.
Thus, there may be potential benefits if we take advantage of
the per-frame labeling task.

Instead of focusing on learning subtle difference near start points, we propose an alternative framework, \ie startNet, and address ODAS in two stages: classification (using ClsNet) and localization (using LocNet). ClsNet conducts per-frame labeling as an auxiliary task based on the spatial-temporal feature aggregation from input videos, and generates score distributions of action classes as a high-level representation. Based on the historical trend of score distributions, LocNet predicts class-agnostic start probability at each time (see Fig~\ref{fig:idea} (b)). At the end, late fusion is applied on the outputs of both modules to generate the final result. When designing LocNet, we consider the implicit temporal constraint between action starts-- two start points are unlikely to be close by. To impose the temporal constraint into the framework under the online setting, historical decisions are taken into account for later predictions. To optimize the long-term reward for start detection, LocNet is trained using reinforcement learning techniques. The proposed framework and its variants are validated on THUMOS'14~\cite{THUMOS14} and ActivityNet~\cite{caba2015activitynet}. Experimental results show that our approach significantly outperforms the state-of-the-art by $10\%$-$30\%$ p-mAP under offsets of $1$-$10$ seconds on THUMOS'14, and achieves comparable p-mAP with $10$ times smaller time offset on ActivityNet. 

%% file: related_work.tex
\section{Related Work}
\label{sec:related_work}
 \vspace{-5pt}
\xhdr{Temporal Action Detection}.
Most existing methods~\cite{shou2016temporal,zhao2017temporal,dai2017temporal,buch2017sst,gao2017turn,chao2018rethinking} on temporal action detection formulate the problem in an offline manner.
These methods segment actions from long, untrimmed videos and require observing the entire video before making a decision. S-CNN~\cite{shou2016temporal} localizes actions with three stages: action proposal generation, proposal
classification, and proposal regression. Dai~\etal~\cite{dai2017temporal} proposed TCN which incorporates
local context of each proposal for proposal ranking. By sharing features between proposal generation and classification, R-C3D~\cite{xu2017r} reduces computational cost significantly. Buch~\etal~\cite{buch2017sst}~propose an efficient proposal generation model that avoids working on overlapping regions. Instead of treating temporal action detection as segment-level classification, Shou~\etal~\cite{shou2017cdc} propose CDC network to produce per-frame predictions using 3D convolutional networks.

\xhdr{Online Action Detection}. Online action detection is usually solved as a per-frame labeling task~\cite{de2016online} on live, streaming videos. As soon as a video frame arrives, it is classified to an action class or background without accessing future frames. De Geest~\etal~\cite{de2016online} first introduced the problem and proposed several models as baselines. Gao~\etal~\cite{gao2017red} propose a Reinforced Encoder-Decoder network for action anticipation and treat online action detection as a special case of their framework. Temporal Recurrent Networks~\cite{xu2018temporal} set a new state-of-the-art performance by conducting current and future action detection jointly. With the same goal of online per-frame labeling, these methods can serve as ClsNet in our framework. 

\xhdr{Early Action Detection}. Early action detectors detect actions after only processing a fraction of videos. The earlier a detector recognizes an action, the better it performs. Hoai~\etal~\cite{hoai2014max} solve this problem by proposing a max-margin framework with structured SVMs. However, this method works on simple scenarios, \eg, one video contains only one action. Ma~\etal~\cite{ma2016learning} design a ranking loss for training assuming that the gaps of predicted scores between correct and incorrect actions should be non-decreasing when an model observes more of an activity.

\xhdr{Online Detection of Action Start (ODAS)}. As with early action detection, ODAS also aims to recognize actions as soon as possible. Specifically, it focuses on detecting action starts and tries to minimize the time delay of identifying the start point of an action. To the best of our knowledge,~\cite{shou2018online} is the first and only work that is designed to address ODAS. They solve the problem by encouraging a classification network to learn a representation that can separate action starts from their preceding backgrounds. To achieve the goal, they force the learned representation of an action start window to be similar to that of the following action window and different from that of the preceding background.

\xhdr{Sequential Search with RL}. Reinforcement learning (RL) techniques are popular for sequential search problems, since RL allows models to be optimized for long-term rewards. Caicedo~\etal~\cite{caicedo2015active} propose a framework based on Deep Q-learning~\cite{mnih2015human} that transforms an initial bounding box iteratively until it lands on an object. In order to speed up object detection on large images, Gao~\etal~\cite{gao2018dynamic} design a coarse-to-fine framework also based on Deep Q-learning that sequentially selects regions to zoom in only when it is needed. Wu~\etal~\cite{wu2018blockdrop} propose BlockDrop that trains with policy gradient~\cite{sutton2018reinforcement} and improved computational efficiency by dropping unnecessary blocks of ResNets~\cite{he2016deep}. AdaFrame~\cite{wu2018adaframe} is also optimized with policy gradient to reduce computations of LSTM by skipping input frames.

%% file: method.tex
\section{Action Start Detection Network (StartNet)}
\label{sec: method}
\begin{figure*}[t]
    \begin{center}
        \includegraphics[width=0.9\linewidth]{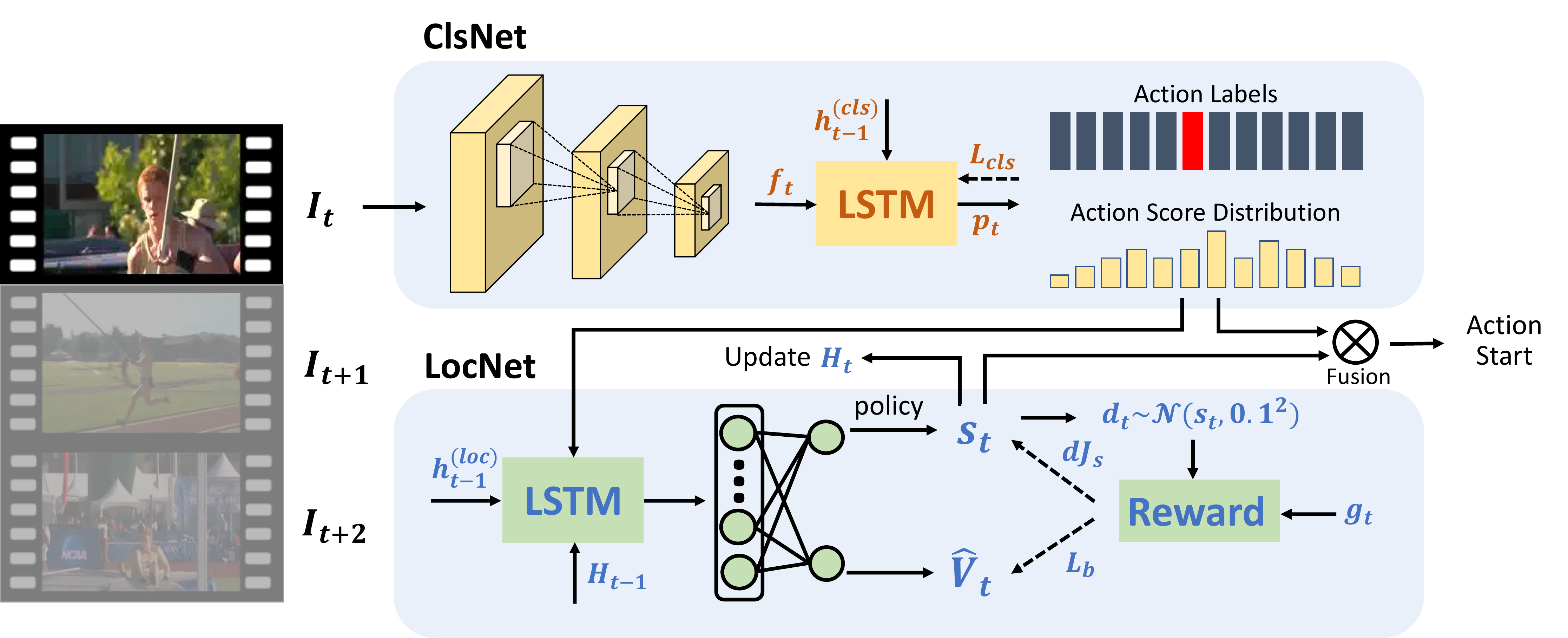}
    \end{center}
    \vspace{-5pt}
    \caption{
Our method works in two stages with ClsNet and LocNet. ClsNet: at time $t$, features, $\textbf{f}_t$, are extracted by deep convolutional networks and input to an one-layer LSTM; The LSTM generates action score distributions at each time step and ClsNet is optimized with cross-entropy loss between action labels and the generated action scores. LocNet: after action score generation, it inputs together with a historical decision vector, $\textbf{H}$, to a second one-layer LSTM which works as an agent to generate two-dimensional start probability sequentially; $\textbf{H}$ is updated and the state is changed  accordingly; The agent is trained using policy gradient mechanism to optimize long-term reward of start localization. At the end, results from ClsNet and LocNet are fused to obtain the final action start detection results at each time step. Here, ClsNet is implemented with LSTM. CNN and C3D can also be used to construct ClsNet (see Sec.~\ref{sec: clsNet} for details).
    }
    \label{fig:framework}
    \vspace{-10pt}
\end{figure*}

The input of an ODAS system is untrimmed, streaming video frames $\{I_1, I_2, ..., I_t\}$.
The system processes each video frame sequentially and detects the start
of each action instance. At time step $t$, it outputs a probability
distribution, $\textbf{as}_t^k$, which indicates
the start probability of the action class $k$, without accessing any future information. 

The overview of the proposed framework is illustrated in Fig.~\ref{fig:framework}. The framework contains two sub-networks,~\ie, a classification network (ClsNet) and a localization network (LocNet). ClsNet focuses on per-frame class labeling. It takes the raw video frames as input and outputs action class probabilities at every time step in an online manner. ClsNet serves two purposes. First, it learns simpler but useful representation for localizing action starts. Second, the classification results can be combined later with the localization results to produce the action starts for each class. LocNet takes the output of ClsNet together with the historical decision vector as inputs. At each time step, it outputs a two-dimensional probability distribution indicating the probability that this frame contains an action start. The historical decision vector records its predictions in the previous $n$ steps in order to model the effect of historical decisions on later ones. Finally, the results of the two networks are fused to construct the final output.

\subsection{Classification Network (ClsNet)}
\label{sec: clsNet}
Inspired by recent online action detection methods~\cite{de2016online,gao2017red,xu2018temporal}, we utilize recurrent networks, specifically, LSTM~\cite{hochreiter1997long}, to construct ClsNet. At each time $t$, it uses the previous hidden state $\textbf{h}^{(cls)}_{t-1}$, the cell $\textbf{c}^{(cls)}_{t-1}$, and the feature, $\textbf{f}_{t}$, extracted from the current video frame, $I_{t}$, as inputs, to update its hidden state $\textbf{h}^{(cls)}_{t}$ and cell $\textbf{c}^{(cls)}_{t}$.
Then, the likelihood distribution over all the action classes can be obtained in Eq.~\ref{eq: cls_cls},
\begin{equation}
\label{eq: cls_cls}
\textbf{p}_{t} = softmax(\textbf{W}_{cls}^{T}\textbf{h}^{(cls)}_{t} + \textbf{b}),
\end{equation}
where $\textbf{p}_{t}$ is a $K$ dimensional vector and $K$ indicates the number of action classes including background.

To learn ClsNet, action class label for each frame is needed. The cross-entropy loss, $L_{cls}(\textbf{W}_c)$, is used for optimization during training, where $\textbf{W}_c$ represents the parameter set of ClsNet.

We observe that ClsNet can be implemented with different architectures. Thus, we validate our framework using two additional structures as the backbone of ClsNet, \ie, CNN and C3D~\cite{tran2015learning}. CNN conducts action classification based only on the arriving frame, $I_t$. It focuses on the spatial information of the current frame without considering temporal patterns of actions. C3D labels $I_t$ based on each temporal segment consisting of 16 consecutive video frames, from $I_{t-15}$ to $I_t$. It captures spatial and temporal information jointly using 3D convolutional operations. Comparisons and explanations are discussed in Sec.~\ref{sec: experiments}.

\subsection{Localization Network (LocNet)}
As discussed in Sec.~\ref{sec: intro}, historical action scores can provide useful cues for identifying action starts. At time $t$, LocNet observes the action score distribution over classes of each frame, $\textbf{p}_{t}$, obtained from ClsNet and outputs a two-dimensional vector, $\textbf{s}_{t}$, indicating the start and non-start probability distribution. 

The start probability is generated sequentially. In general, if an action starts at time step $t$, there is a low probability that  another action also starts at time $t+1$, given reasonable frames per second (FPS). Thus, there are implicit temporal constraints between nearby start points. To enable the model to consider constraints between decisions, we record the historical decisions made by LocNet and use the history to influence later decisions. To enable long-term decision planning, we formulate the problem as a Markov Decision Process (MDP) and use reinforcement learning to optimize our model. When making a decision\footnote{The term ``action" is generally used in reinforcement learning, we use ``decision" instead to remove the confusion with action class.}, the model not only considers the effect of the decision at the current step, but also how it will influence the later ones by maximizing the expected long-term reward. In the following, we first discuss the inference phase of LocNet and then the training phase in detail.
\vspace{-5pt}
\subsubsection{Inference Phase}
LocNet is built upon a LSTM structure. It acts as an agent which interacts with historical action scores recurrently. During testing, at each state, the agent makes a decision (predicts start probability) that produces the maximum expected long-term reward and updates the state according to the decision. To model the dependency between decisions, we incorporate the record of historical decisions (the decisions made by the agent at previous steps) as a part of the state. The state update procedure is described in Eq.~\ref{eq: lstm_loc} and~\ref{eq: cls_loc},
where $\textbf{H}_{t-1}=\textbf{s}_{t-n:t-1}$ indicates  historical decisions from step $t-n$ to $t-1$ and $[\textbf{p}_{t}, \textbf{H}_{t-1}]$ indicates the concatenation of the vectors. At the beginning, $\textbf{H}$ is initialized with zeros.
\begin{equation}
\label{eq: lstm_loc}
\textbf{h}^{(loc)}_{t}, \textbf{c}^{(loc)}_{t} = LSTM(\textbf{h}^{(loc)}_{t-1}, \textbf{c}^{(loc)}_{t-1}, [\textbf{p}_{t}, \textbf{H}_{t-1}]).
\end{equation}
\begin{equation}
\label{eq: cls_loc}
\textbf{s}_{t} = softmax(\textbf{W}_{loc}^{T}\textbf{h}^{(loc)}_{t} + \textbf{b}).
\end{equation}
\subsubsection{Training Phase}
We train an agent that acts optimally based on the state of the environment. The goal is to maximize the reward by changing the predicted start probability distribution: at a given state, the start probability should be increased when the decision introduces bigger reward and be decreased otherwise. The start prediction procedure is formulated as a decision making policy defined using Gaussian distribution. Following~\cite{mnih2014recurrent, wu2018adaframe}, the policy is trained by optimizing with $d_t$, where $d_{t}$, is sampled from $\pi(.|\textbf{h}^{(loc)}_{t}, \textbf{p}_{t}, \textbf{H}_{t-1}) =\mathcal{N}(s_{t}, 0.1^2)$ and $s_{t}$ indicates the output start probability that determines the Gaussian distribution. 

\xhdr{Reward function}. Each decision at a given state is associated with an immediate reward to measure the decision made by the agent at the current time. With the goal of localizing start points, we define the immediate reward function in Eq.~\ref{eq: reward}, where $g_t \in \{0,1\}$ indicates the ground-truth label of action start and $d_t$ is the sampled start probability. The reward function encourages a high probability when there is an actual start and a low probability when there is not by giving a negative reward. Considering the sample imbalance between start points and background, weighted rewards are used by setting a parameter $\alpha$. In particular, we set $\alpha$ to be the ratio between the number of negative samples to positive samples for each dataset. 
\begin{equation}
\label{eq: reward}
r_t =\alpha g_td_t-(1-g_t)d_t.
\end{equation}

The long-term reward is the summation of discounted future rewards. In order to maximize the expected long-term reward, the policy is trained by maximizing  the objective in Eq.~\ref{eq: objective}, where $\textbf{W}_s$ represents the parameters of the network and $\gamma$ is a constant scalar for calculating the discounted  rewards over time. 
\begin{equation}
\label{eq: objective}
J_s(\textbf{W}_{s}) = \mathop{\mathbb{E}}_{d_{t} \sim \pi(.|\textbf{W}_{s})}[\sum_{i=0} \gamma^{i} r_{t+i}].
\end{equation}

\xhdr{Optimization}. When optimizing Eq.~\ref{eq: objective}, it is not possible to train the network using error back propagation directly, since the objective is not differentiable. Following~\cite{sutton2018reinforcement}, we use policy gradient to calculate the expected gradient of $J_s$ as in Eq.~\ref{eq: pg}, where $R_t = \sum_{i=0} \gamma^{i} r_{t+i}$ indicates the long-term reward at time step $t$ and $\hat{V}_t$ is a baseline value which is widely used in policy gradient frameworks to reduce the variance of the gradient. The principle of policy gradient is to maximize the probability of an action with high reward given a state. The baseline value encourages that the model is optimized in the direction of performance improvement.
\begin{equation}
\label{eq: pg}
\bigtriangledown_{\textbf{W}_{s}} J_s = \mathop{\mathbb{E}}[\sum_{t=0}^{\infty }(R_{t}-\hat{V}_t)\bigtriangledown_{\textbf{W}_{s}}log\pi(.|\textbf{W}_{s})].
\end{equation}

Following~\cite{wu2018adaframe}, we use the expected long-term reward at the current state as the baseline value and approximate it by minimizing the $l_2$ loss: $L_{b}(\textbf{W}_{b}) = \frac{1}{2}||R_t - \hat{V}_t||_2$. The training procedure of LocNet is summarized in Alg.~\ref{alg: train}.
\begin{algorithm}
  \caption{Training Process of LocNet}\label{alg: train}
  \small
  \begin{algorithmic}
      \State Initialize parameters, $\textbf{W}_s$, of LocNet
      \For{ iteration = 1:$N$}
        \State Obtain training sequence samples of length $T_{loc}$
        \For{t = 1:$T_{loc}$}
          \State Obtain $s_t$ based on current policy
          \State Sample decisions: $d_{t} \sim \mathcal{N}(s_{t}, 0.1^2)$
          \State Obtain $r_t$ and $\hat{V}_t$ for each sample
        \EndFor
        \State Compute $R_{1:T_{loc}}$, $\bigtriangledown_{\textbf{W}_{s}} J_s$ and $L_b(\textbf{W}_b)$
        \State Update parameters, $\textbf{W}_s$, of LocNet
      \EndFor
  \end{algorithmic}
\end{algorithm}

The full objective including the loss term in ClsNet is shown in Eq.~\ref{eq: full_objective}, where $\lambda_1$ and $\lambda_2$ are constant scalars. 
\begin{equation}
\label{eq: full_objective}
\min L_{cls}(\textbf{W}_{c})+\lambda_1L_{b}(\textbf{W}_{b})- \lambda_2 J_s(\textbf{W}_{s}).
\end{equation}

\subsection{Late Fusion}
\label{sec: fusion}
ClsNet outputs an action score distribution and LocNet produces class-agnostic start probabilities at each time step. Then, late fusion is applied to obtain the start probability for each action class, $\textbf{as}_t^{k}$, following Eq.~\ref{eq: fusion}, where superscript ${1:K-1}$ indicates positive action classes and $0$ indicates background.
\begin{equation}
\label{eq: fusion}
\textbf{as}_t^{k} =\left\{\begin{matrix}
 s_t\textbf{p}_t^{1:K-1} & k=1:K-1\\ 
 (1-s_t)\textbf{p}_t^{0}& k=0
\end{matrix}\right..
\end{equation}

\xhdr{Action start generation}. Follow~\cite{shou2018online}, final action starts are generated online if all of the three conditions are satisfied: (i) $c_t = \underset{k}{\mathrm{argmax}} (\textbf{as}_t^{k})$ is an action; (ii) $c_t \neq c_{t-1}$ and (iii) $\textbf{as}_t^{c_t}$ exceeds a threshold. We set this threshold to 0 by default. An action score sequence generated by ClsNet can also generate action start points online following this procedure. LocNet can locally adjust the start point by boosting time points with higher start probabilities and suppressing those with lower start probabilities.

%% file: experiments.tex
\section{Experiments}
\label{sec: experiments}
To validate the proposed framework, we conduct extensive experiments
on two large-scale action recognition datasets, \ie, THUMOS'14~\cite{THUMOS14}
and ActivityNet v1.3~\cite{caba2015activitynet}.

\xhdr{Evaluation protocol}. To permit fair comparisons, we use the
point-level average precision (p-AP) proposed in~\cite{shou2018online}
to evaluate our framework. Under this protocol, each action
start prediction is associated with a time point. For each action
class, predictions of all frames are first sorted in descending order
based on their confidence scores and then measured accordingly. An
action start prediction is counted as correct only if it matches the
correct action class and its temporal distance from a ground-truth
point is smaller than an offset threshold (offset tolerance). Similar
to segment-level average precision, no duplicate detections are allowed
for the same ground-truth point. p-mAP is then calculated by averaging
p-AP over all the action classes. 

Following~\cite{shou2018online}, we
use two metrics based on p-AP to evaluate our framework on THUMOS'14.
First, we use p-AP under different offset tolerances, varying from $1$
to $10$ seconds. Also, we adopt the metric \emph{AP depth at recall (Rec) X$\%$}
which averages p-AP on the Precision-Recall curve with the recall
rate from $0\%$ to X$\%$. p-mAPs under different offset thresholds are
then averaged to obtain the final average p-mAP at each depth. This
metric is particularly used to evaluate top ranked predictions and to
measure what precision a system can achieve if low recall is allowed.
For ActivityNet, we evaluate our methods using p-mAP under offset
thresholds of $1$-$10$ seconds at depth $Rec$=$1.0$.

\xhdr{Baselines}. We compare the proposed framework with the
state-of-the-art method, \ie, \emph{Shou et al.}~\cite{shou2018online}
and two baselines that were presented in~\cite{shou2018online}, \ie, \emph{SceneDetect} and \emph{ShotDetect}. The numbers were obtained from the authors~\cite{shou2018online}. Comparison results with \emph{Shou et al.}~\cite{shou2018online} demonstrate the superior performance of StartNet. \emph{SceneDetect} and \emph{ShotDetect} are also two-stage methods. Similar to two-stage frameworks of object detection, they first conduct localization by getting action start proposals, which are generated by soft boundary detectors, and then classify them to different classes. Comparison with \emph{SceneDetect} and \emph{ShotDetect} shows the effectiveness of our decomposition design. Our framework trained by policy gradient is indicated by~\emph{StartNet-PG}.

\begin{table*}[]
    \centering
    \footnotesize
    \begin{tabular}{l|l||c|c|c|c|c|c|c|c|c|c}
        & Offsets (second) & 1 & 2 & 3 & 4 & 5 & 6 & 7 & 8 & 9 & 10 \\
        \midrule
        \multirow{3}{*}{Baselines} & SceneDetect~\cite{Scenedetect} &1.0& 2.0& 2.3& 3.1& 3.6& 4.1& 4.7& 5.0& 5.1& 5.2 \\
        & ShotDetect~\cite{Shotdetect} &1.1& 1.9& 2.3& 3.0& 3.4& 3.9& 4.3& 4.5& 4.6& 4.9  \\
        & \emph{Shou et al.}~\cite{shou2018online}  &3.1& 4.3& 4.7& 5.4& 5.8& 6.1& 6.5& 7.2& 7.6& 8.2 \\
        \midrule
        \multirow{3}{*}{StartNet-PG} & C3D~\cite{tran2015learning} + LocNet & 6.8& 8.0& 9.4& 10.1& 10.6& 10.9& 10.9& 11.1& 11.2& 11.2 \\
        & CNN~\cite{wang2016temporal} + LocNet &17.0& 23.6& 27.6& 29.9& 31.3& 32.1& 33.2& 33.5& 33.9& 34.5  \\
        & \textbf{LSTM}~\cite{hochreiter1997long} + \textbf{LocNet} &\textbf{19.5} &\textbf{27.2}& \textbf{30.8}&\textbf{33.9}&\textbf{36.5}&\textbf{37.5}&\textbf{38.3}&\textbf{38.8}&\textbf{39.5}&\textbf{39.8}
    \end{tabular}
    \caption{Comparisons using p-mAP at depth $Rec$=$1.0$ on THUMOS'14. Results are under different offset thresholds. ClsNet is implemented with different structures, \ie, C3D, CNN and LSTM. CNN and LSTM are using TS features.}
    \label{tab: thumos_soa}
\end{table*}

\begin{table*}[]
    \centering
    \footnotesize
    \begin{tabular}{l|l||c|c|c|c|c|c|c|c|c|c}
        & Depth Rec. & @0.1 & @0.2 & @0.3 & @0.4 & @0.5 & @0.6 & @0.7 & @0.8 & @0.9 & @1.0 \\
        \midrule
        \multirow{3}{*}{Baselines} & SceneDetect~\cite{Scenedetect} &30.0& 18.3& 12.2& 9.1& 7.2& 6.1& 5.2& 4.6& 4.0&3.6  \\
        & ShotDetect~\cite{Shotdetect}  &26.3& 15.9& 11.3& 8.6& 6.8& 5.8& 4.9& 4.3& 3.8& 3.4\\
        & \emph{Shou et al.}~\cite{shou2018online} &42.7& 27.3& 19.8& 14.9& 11.8& 10.0& 8.5& 7.4& 6.6& 5.9  \\
        \midrule
        \multirow{3}{*}{StartNet-PG} 
        & C3D~\cite{tran2015learning} + LocNet &34.8& 27.7& 22.6& 19.0& 16.3& 14.4& 12.9& 11.8& 10.8& 10.0   \\
        & CNN~\cite{wang2016temporal} + LocNet  &71.8& 64.7& 58.0& 52.4& 47.2& 43.3& 39.5& 35.9& 32.5& 29.6 \\
        & \textbf{LSTM}~\cite{hochreiter1997long} + \textbf{LocNet} &\textbf{77.4}& \textbf{70.2}& \textbf{64.5}& \textbf{59.1}& \textbf{54.2}& \textbf{49.3}& \textbf{45.1}& \textbf{41.2}& \textbf{37.6}&\textbf{34.2}
    \end{tabular}
    \caption{Comparisons using average p-mAP at different depths on THUMOS'14. Average p-mAP means averaging p-mAP over offsets from $1$ to $10$ seconds. ClsNet is implemented with different structures, \ie, C3D, CNN and LSTM. CNN and LSTM are using TS features.}
    \label{tab: thumos_soa_depth}
     \vspace{-10pt}
\end{table*}

\xhdr{Implementation details}. Following~\cite{xu2018temporal,gao2017red, shou2018online}, decisions are made on short temporal chunks, $\mathcal{C}_t$, where $I_t$ is its central frame. The appearance feature (RGB) of $\mathcal{C}_t$ is extracted from $I_t$ and the motion feature (optical flow) is computed using the whole chunk as input. Following~\cite{xu2018temporal,gao2017red}, chunk size is fixed to 6 and image frames are obtained at 24 FPS. Two adjacent chunks are not overlapping, thus, there are exactly 4 chunks per second. Following~\cite{xu2018temporal}, for ClsNet, we set the size of LSTM's hidden state to $4096$ and the length of each training sequence to 64.  When using CNN, we finetune an fully-connected (FC) layer with different CNN features as input (see feature descriptions for each dataset). C3D is pretrained on Sports-1M~\cite{karpathy2014large} and finetuned for the per-frame labeling task on each dataset. Hidden state of LocNet is set to $128$ and the length of each training sequence, $T_{loc}$, is fixed to $16$. Following~\cite{wu2018adaframe}, $\gamma$ in Eq.~\ref{eq: objective} is fixed to $0.9$. The length of the historical decision vector, $n$, is set to 8. $\lambda_1$ and $\lambda_2$ in Eq.~\ref{eq: full_objective} are fixed to $1$. We adopt an alternating strategy for classification and localization training: ClsNet is first trained and fixed afterwards, and then LocNet is trained upon the pre-trained ClsNet. We implement the models in PyTorch~\cite{pytorch}, and set batch size to 32 for THUMOS'14 and 64 for ActivityNet. For parameter optimization, we used the Adam~\cite{kingma2014adam} optimizer with learning rate $5e^{-4}$ and weight decay $5e^{-4}$. 

\subsection{Experiments on THUMOS'14}
\xhdr{Dataset}. THUMOS'14~\cite{THUMOS14} is a popular benchmark for temporal action detection. It contains 20 action classes related to sports. There are only trimmed videos in the training set which makes it not appropriate for training ODAS methods. Following~\cite{shou2018online}, we use the validation set (including 200 untrimmed videos, 3K action instances) for training and the test set (including 213 untrimmed videos, 3.3K action instances) for testing.

\xhdr{Feature description}. Two types of features are adopted on THUMOS'14 dataset, \emph{RGB} and \emph{Two-Stream (TS)} features. Following~\cite{gao2017red, xu2018temporal}, we extract appearance (RGB) feature at the \emph{Flatten 673} layer of ResNet-200~\cite{he2016deep} and motion feature at the \emph{global pool} layer of BN-Inception~\cite{ioffe2015batch} with optical flows of $6$ consecutive frames as inputs. The TS feature is the concatenation of appearance and motion features, which are extracted with models\footnote{https://github.com/yjxiong/anet2016-cuhk.} pre-trained on ActivityNet.

\begin{table*}[]
    \centering
    \footnotesize
    \begin{tabular}{l|l||c|c|c|c|c|c|c|c|c|c}
        Features & Offsets (second) & 1 & 2 & 3 & 4 & 5 & 6 & 7 & 8 & 9 & 10 \\
        \midrule
        \multirow{3}{*}{RGB} & ClsNet-only &11.8  &17.2  &21.3  &24.9  & 27.9 &28.7  &29.5  &30.0  &30.4  & 30.7 \\
        & StartNet-CE &13.7  &20.7  &23.8  &27.2  &29.4  &30.7  &31.9  &32.5  & 33.2 &33.6  \\
        & \textbf{StartNet-PG} &\textbf{15.9}  &\textbf{21.0}  &\textbf{24.8}  &\textbf{28.4}  &\textbf{30.7}  &\textbf{31.8}  &\textbf{33.0}  &  \textbf{33.5}&\textbf{34.0}  & \textbf{34.4} \\
        \midrule
        \multirow{3}{*}{Two Stream} & ClsNet-only & 13.9 &21.6  &25.8  &28.9  &31.1  &32.5  &33.5  &34.3  &34.8  &35.2  \\
        & StartNet-CE &17.4  &25.4  &29.8  &33.0  &34.6  &36.3  &37.2  &37.7  & 38.6 &38.8  \\
        & \textbf{StartNet-PG} &\textbf{19.5} &\textbf{27.2}& \textbf{30.8}&\textbf{33.9}&\textbf{36.5}&\textbf{37.5}&\textbf{38.3}&\textbf{38.8}&\textbf{39.5}&\textbf{39.8}
    \end{tabular}
    \caption{Ablation study of our framework using p-mAP at depth $Rec$=$1.0$ on THUMOS'14. LSTM is used to implement ClsNet. Different offset thresholds are used to evaluate our framework with different features. Best performance is marked in bold. }
    \label{tab: thumos_ablation}
\end{table*}

\begin{table*}[]
    \centering
    \footnotesize
    \begin{tabular}{l|l||c|c|c|c|c|c|c|c|c|c}
        Features & Depth Rec.& @0.1 & @0.2 & @0.3 & @0.4 & @0.5 & @0.6 & @0.7 & @0.8 & @0.9 & @1.0 \\
        \midrule
        \multirow{3}{*}{RGB} & ClsNet-only &71.2& 61.1& 52.8& 47.0& 42.0& 37.7&34.0& 30.6& 27.5& 25.3 \\
        & StartNet-CE & 73.2& 64.5& 56.8& 50.2& 45.1& 40.5& 36.6& 33.5& 30.5& 27.7\\
        & \textbf{StartNet-PG}  & \textbf{73.6}& \textbf{65.0}& \textbf{58.0}& \textbf{51.2}& \textbf{45.9}& \textbf{41.5}& \textbf{37.8}& \textbf{34.3}& \textbf{31.5}& \textbf{28.8}  \\
        \midrule
        \multirow{3}{*}{Two Stream} & ClsNet-only &71.3&63.0& 56.9& 52.0& 46.9& 42.3& 38.7& 35.0& 31.8& 29.2  \\
        & StartNet-CE & 72.7 &65.6& 60.2& 55.3& 51.0& 46.8& 43.0& 39.2 &36.0 &32.9  \\
        & \textbf{StartNet-PG} &\textbf{77.4}& \textbf{70.2}& \textbf{64.5}& \textbf{59.1}& \textbf{54.2}& \textbf{49.3}& \textbf{45.1}& \textbf{41.2}& \textbf{37.6}&\textbf{34.2}
    \end{tabular}
    \caption{Ablation study of our framework using average p-mAP at different depths on THUMOS'14. At each depth, we average p-mAP over offset thresholds from $1$ to $10$ seconds. LSTM is used to implement ClsNet. Best performance is marked in bold. }
    \label{tab: thumos_ablation_depth}
    \vspace{-10pt}
\end{table*}

\begin{table*}[]
    \centering
    \footnotesize
    \begin{tabular}{l|l|c|c|c|c|c|c|c|c|c|c}
        &Offsets (second) & 1 & 2 & 3 &4&5&6&7&8&9 & 10  \\
        \midrule
        \multirow{4}{*}{Baselines} & SceneDetect~\cite{Scenedetect} & -- & -- & -- & -- & -- & -- & -- & -- & -- & 4.7  \\
        &ShotDetect~\cite{Shotdetect} & -- & -- & -- & -- & -- & -- & -- & -- & -- &6.1 \\
        &\emph{Shou et al.}~\cite{shou2018online} & -- & -- & -- & -- & -- & -- & -- & -- & -- & 8.3 \\
        \midrule
        \midrule
        \multirow{6}{*}{StartNet} & ClsNet-only-VGG &2.7 &4.1& 5.1& 5.9&6.7&7.5& 8.1&8.7&9.2&9.8 \\
        &StartNet-CE-VGG&4.2  &6.1&7.4&8.7&9.7&10.5&11.4&12.0 &12.6&13.1  \\
        &\textbf{StartNet-PG-VGG} &\textbf{6.0}  & \textbf{7.6}& \textbf{8.8}& \textbf{9.8}& \textbf{10.7}& \textbf{11.5}& \textbf{12.2}& \textbf{12.6}& \textbf{13.1} &\textbf{13.5}  \\
        \cmidrule{2-12}
&ClsNet-only-TS&4.2& 6.1& 7.7& 8.8& 9.8& 10.7& 11.3& 12.2& 13.0&13.6\\
        &StartNet-CE-TS &6.0&8.3&10.1&11.7& 12.9& 13.9& 15.0& 15.8& 16.7& 17.5  \\
        &\textbf{StartNet-PG-TS} &\textbf{8.1}&\textbf{10.2} & \textbf{11.8}& \textbf{13.3}& \textbf{14.4}& \textbf{15.3}& \textbf{16.1}& \textbf{16.7}& \textbf{17.4} & \textbf{18.0}
    \end{tabular}
    \caption{Comparisons using p-mAP under varing offset thresholds at depth $Rec$=$1.0$ on ActivityNet. ClsNet is implemented with LSTM. Numbers of baseline methods are cited from~\cite{shou2018online}. -- indicates that numbers are not provided in~\cite{shou2018online}.}
    \label{tab: act_soa}
       \vspace{-10pt}
\end{table*}
\vspace{-10pt}
\subsubsection{Evaluation Results}
\vspace{-5pt}
Comparisons with previous methods are shown in Table~\ref{tab: thumos_soa} and Table~\ref{tab: thumos_soa_depth}. Table~\ref{tab: thumos_soa} shows comparisons based on p-mAP at depth $Rec$=$1.0$ under different offset thresholds. All previous methods are under $4\%$ p-mAP at 1 second offset, while StartNet with LSTM achieves $19.5\%$ p-mAP, outperforming the state-of-the-arts largely by over $15\%$. At $10$ seconds offset, previous methods obtain less than $9\%$ p-mAP and StartNet (LSTM) improves over \emph{Shou et al.}~\cite{shou2018online} by $30\%$ p-mAP. Table~\ref{tab: thumos_soa_depth} shows comparisons based on average p-mAP (averaging over offsets from $1$ to $10$ seconds) at different depths. The results demonstrate that StartNet with LSTM outperforms previous methods significantly (by around $30\%$-$20\%$ average p-mAP) at depth from $Rec$=$0.1$ to $Rec$=$1.0$. Obviously, under both metrics, StartNet outperforms previous methods by a very large margin.

\begin{figure}[]
    \begin{center}
        \includegraphics[width=0.9\linewidth]{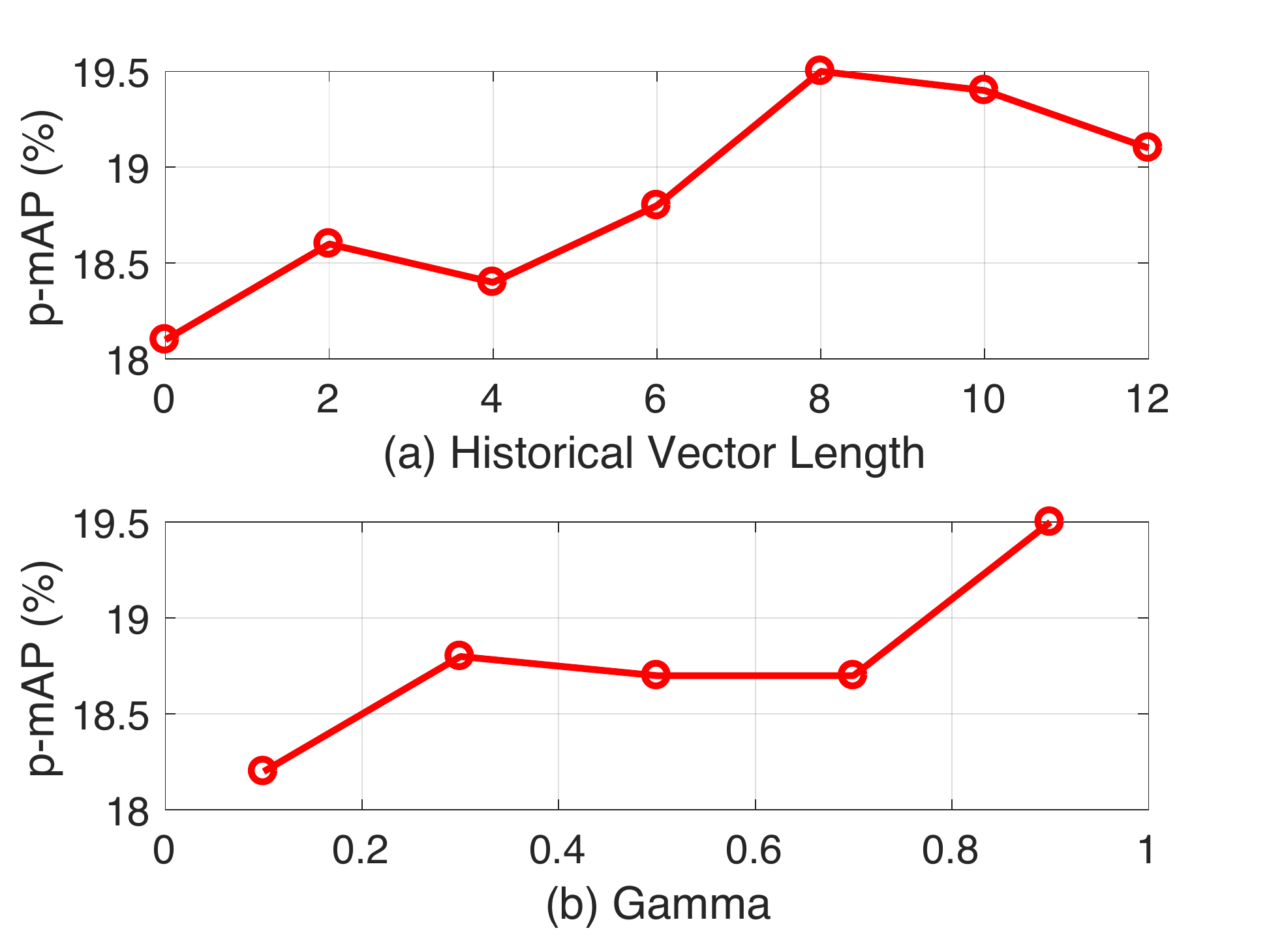}
    \end{center}
    \vspace{-10pt}
    \caption{
        Ablation study of LocNet: (a) effect of length of historical decision vector (b) effect of different gamma values in Eq.~\ref{eq: objective}.
        Generally, the model performs better with bigger gamma and longer historical decision vector.
    }
    \label{fig: ablation}
       \vspace{-10pt}
\end{figure}
\begin{figure}[]
    \begin{center}
        \includegraphics[width=1.0\linewidth]{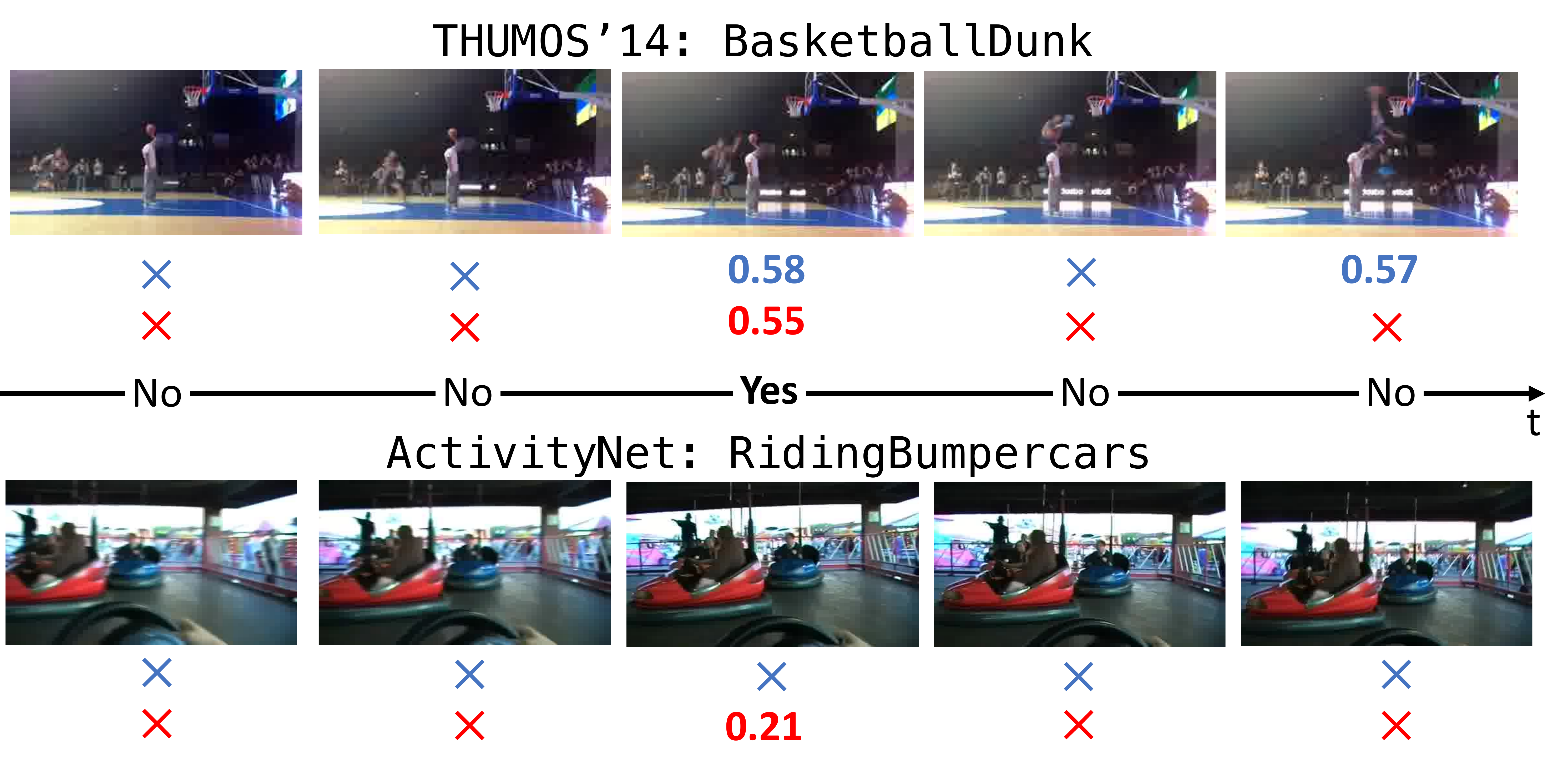}
    \end{center}
    \vspace{-10pt}
    \caption{
        Qualitative results on THUMOS'14 and ActivityNet after action start generation (see Sec.~\ref{sec: fusion}). $\times$ means no starts are detected at those times. Numbers indicate the scores of detected action starts. Results of {\color{blue}{ClsNet}} and {\color{red}{StartNet}} are marked in blue and red, respectively. Yes/No (ground-truth) indicates if an action of the associated class starts at the time. Best viewed in color.
    }
    \label{fig: qualitative}
           \vspace{-10pt}
\end{figure}
\vspace{-10pt}
\subsubsection{Ablation Experiments}
\label{sec: ablation_thumos}
\vspace{-5pt}
\xhdr{ClsNet implemented with different structures}. Comparisons among StartNet with different ClsNet's backbones are shown in Table~\ref{tab: thumos_soa} and Table~\ref{tab: thumos_soa_depth}. \emph{LSTM+LocNet} achieves the best performance among the three structures. It is worth noticing that \emph{C3D} performs much worse than \emph{CNN} and \emph{LSTM}, which shows its disadvantage in the online action detection task. In offline setting, \emph{C3D} can observe the entire temporal context of an action before making a decision, but it has to recognize the occurring action based only on the preceding temporal segment when working online. Compared to \emph{LSTM}, it has no recurrent structure to learn long-term patterns. Compared to \emph{CNN}, it has more complicated operations and is more prone to overfitting. \emph{Shou et al.}~\cite{shou2018online} chose C3D as its backbone and proposed sophisticated training strategies for optimization. However, C3D may not be suitable for the task according to our comparisons with other structures. Even with \emph{C3D}, StartNet still significantly outperforms \emph{Shou et al.}~\cite{shou2018online}, which demonstrates the effectiveness of our framework.  Since \emph{LSTM+LocNet} achieves the best performance, the following ablation studies are conducted using ClsNet implemented with LSTM.

\xhdr{Effectiveness of LocNet}. The results from ClsNet alone can be used to generate action starts by following the \emph{action start generation procedure} in Sec.~\ref{sec: fusion}. To evaluate the contribution of LocNet, we construct \emph{ClsNet-only} by removing LocNet from our framework.  Results of \emph{ClsNet-only} can also demonstrate the performance of OAD methods if applied on the ODAS task directly. As shown in Table~\ref{tab: thumos_ablation}, \emph{ClsNet-only} has already achieved good results, outperforming \emph{C3D} based methods. When adding \emph{LocNet}, \emph{StartNet-PG} improves \emph{ClsNet-only} by $5\%$-$6\%$ p-mAP with TS feature and by  $4\%$-$5\%$ p-mAP with RGB features under varying offsets. We can also observe a trend that the gaps between \emph{StartNet-PG} and \emph{ClsNet-only} are larger when the offset is smaller. As shown in Table~\ref{tab: thumos_ablation_depth}, \emph{StartNet-PG} outperforms \emph{ClsNet-only} by $5\%$-$6\%$ p-mAP with TS features and about $3\%$-$5\%$ p-mAP with RGB features at different depths. The qualitative comparison in Fig.~\ref{fig: qualitative} shows an example that \emph{ClsNet-only} generate a false positive at the last frame, which may be because that the frame contains a classic appearance of the action, \ie, \emph{Basketball Dunk}. With the help of LocNet, the false positive is corrected by \emph{StartNet-PG}. 

\xhdr{Effectiveness of long-term planning}. In order to investigate the effect of long-term planning, we replace the policy gradient training strategy with simple cross-entropy loss -- $-\beta g_tlog(s_t)-(1-g_t)log(1-s_t)$-- such that every frame is considered independently. This baseline is referred as \emph{StartNet-CE}. Similar to \emph{StartNet-PG}, weight factor, $\beta$, is used to handle sample imbalance. Same as $\alpha$ in Eq.~\ref{eq: reward}, we set $\beta$ equal to the ratio between the number of negative samples and positive ones. As shown in Table~\ref{tab: thumos_ablation} and \ref{tab: thumos_ablation_depth}, \emph{StartNet-PG} significantly outperforms \emph{StartNet-CE} under each offset threshold and at different depths, which proves the usefulness of the long-term planning. 

In order to further investigate effects of parameter settings for LocNet, we conduct an ablation study on different values of the length of historical decision vector, $n$, and gamma in Eq.~\ref{eq: objective} when offset threshold is set to $1$ second and depth $Rec$=$1.0$. Results are shown in Fig.~\ref{fig: ablation}. Increasing the length of the historical decision vector means increasing the dependency of later decisions on previous ones. As is shown, the model performs much better when incorporating historical decisions and it reaches its highest performance when $8$ historical decisions are considered. Increasing gamma indicates increasing the effect of future rewards to the total long-term reward. It shows that when increasing values of gamma, the model performs better. 

\xhdr{Results with different features}. To investigated the performance of our framework when using different features, we add experiments with \emph{ClsNet-only}, \emph{StartNet-CE} and \emph{StartNet-PG} using appearance features (RGB) only. Results are displayed in Table~\ref{tab: thumos_ablation} and Table \ref{tab: thumos_ablation_depth}. We see that when using only RGB features, performance of the three models drops. However, even with RGB features, our method still outperforms \emph{Shou et al.}~\cite{shou2018online} largely.

\xhdr{Effectiveness of two-stage design}. We validate our two-stage design by comparing with \emph{one-stage network} which has similar structure as ClsNet (LSTM) except that we modify it to directly predict action starts for all classes and optimize it with cross-entropy loss. We get $6.5\%$ and $10.2\%$ p-mAP at $1$ second offset (depth $Rec$=$1.0$) using RGB and TS features, respectively. The results are much worse than \emph{StartNet-CE} and \emph{StartNet-PG} (drops about $7\%$ and $9\%$), demonstrating that simply learning classification and localization of action starts jointly is not a good strategy. 

\xhdr{Learning from low-level features}. Our framework uses action score distributions pretrained on an auxiliary task as inputs of LocNet. We believe that learning from this high-level representation is better than learning from low-level noisy features for our task due to the lack of training data. To prove this point, we construct \emph{StartNet-img} where LocNet learns directly from the low-level image features. The p-mAP using RGB and TS features under offsets of $1$ second (depth is $1.0$) is $10.2\%$ and $14.0\%$, respectively, which much under perform our framework (drops about $5\%$). 

\subsection{Experiments on ActivityNet}
\vspace{-5pt}
\xhdr{Dataset}. ActivityNet v1.3~\cite{caba2015activitynet} is one of the largest datasets for action recognition. It contains annotations of 200 action classes. There are around 10K untrimmed videos (15K action instances) in the training set and 5K (7.6K action instances) untrimmed videos in the validation set. Averagely, there are around 1.6 action instances in each video. Following~\cite{shou2018online}, we train our models on the train set and test them on the validation set.

\xhdr{Feature description}. TS feature is constructed by concatenating appearance and motion features that are extracted from TSN model (with BN-Inception)~\cite{wang2016temporal} pretrained on Kinetics~\cite{carreira2017quo}. Besides, we validate our method using appearance features extracted from \emph{fc6} layer of VGG-16~\cite{simonyan2014very}. The VGG-16 model is pretrained on ImageNet~\cite{deng2009imagenet}. VGG-16 features are not as good as ResNet and InceptionNet features for action recognition tasks. We use VGG-16 features to show that our framework can produce reasonable results even when using simple features pretrained only on images.

\xhdr{Training sample strategy of LocNet}. Unlike THUMOS'14 which contains around 16 action instances per video in average, ActivityNet has only one action instance in most of the videos. Thus, ActivityNet has much severer imbalance problem between start and non-start classes. To balance the samples, we randomly select equal numbers of positive and negative sequences for each training batch. Positive sequence is defined as containing at least one action start and negative one contains no action start. Then, $\alpha$ is set to the ratio between the number of negative samples over the number of positive ones after the sample balance.

\xhdr{Evaluation results}. Comparisons of StartNet with previous methods on ActivityNet are shown in Table~\ref{tab: act_soa}. StartNet significantly outperforms previous methods. Specifically, StartNet with TS feature achieves similar performance under $1$ second offset tolerance compared to \emph{Shou et al.}~\cite{shou2018online} under $10$ seconds offset. At offset of $10$ seconds, our method improves \emph{Shou et al.}~\cite{shou2018online} by around $10\%$. It also outperforms \emph{SceneDetect} and \emph{ShotDetect} largely by $13.3\%$ and $11.9\%$, respectively. Even with VGG features pretrained on only images, our method significantly outperforms the state-of-the-arts. Besides, we demostrate the contribution of each module by comparing with \emph{ClsNet-only} and \emph{StartNet-CE} (refer to Sec.~\ref{sec: ablation_thumos} for detailed model description). Results show that by adding LocNet, \emph{StartNet-PG} improves \emph{ClsNet-only} by over $3\%$ (using VGG) and around $4\%$ (using TS) p-mAP. With long-term planning, \emph{StartNet-PG} significantly outperforms \emph{StartNet-CE} under both features, especially when the offset tolerance is small. Qualitative results in Fig.~\ref{fig: qualitative} shows a hard case where \emph{ClsNet-only} misses an action start due to the subtle appearance difference near the start point. With LocNet, \emph{StartNet-PG} successfully captures the start point although the score is low.

%% file: conclusion.tex
\vspace{-12pt}
\section{Conclusion}
\label{sec:conclusion}
\vspace{-3pt}
We proposed StartNet to handle Online Detection of Action Starts. StartNet consists of two networks, \ie, ClsNet and LocNet. ClsNet processes the input streaming video and generates action scores for each video frame. LocNet localizes start points by optimizing long-term planning rewards using policy gradient methods. At the end, results from the two sub-networks are fused to produce the final action start predictions. Experimental results on THUMOS'14 and ActivityNet demonstrate that our framework significantly outperforms the state-of-the-arts. Extensive ablation studies were conducted to show the effectiveness of each module of our method.

%% file: main.bbl
\begin{thebibliography}{10}\itemsep=-1pt

\bibitem{Scenedetect}
\url{https://github.com/Breakthrough/PySceneDetect}.

\bibitem{Shotdetect}
\url{https://github.com/johmathe/Shotdetect}.

\bibitem{pytorch}
\url{http://pytorch.org/}.

\bibitem{buch2017sst}
S.~Buch, V.~Escorcia, C.~Shen, B.~Ghanem, and J.~C. Niebles.
\newblock {SST}: Single-stream temporal action proposals.
\newblock In {\em CVPR}, 2017.

\bibitem{caicedo2015active}
J.~C. Caicedo and S.~Lazebnik.
\newblock Active object localization with deep reinforcement learning.
\newblock In {\em ICCV}, 2015.

\bibitem{carreira2017quo}
J.~Carreira and A.~Zisserman.
\newblock Quo vadis, action recognition? a new model and the kinetics dataset.
\newblock In {\em CVPR}, 2017.

\bibitem{chao2018rethinking}
Y.-W. Chao, S.~Vijayanarasimhan, B.~Seybold, D.~A. Ross, J.~Deng, and
  R.~Sukthankar.
\newblock Rethinking the faster r-cnn architecture for temporal action
  localization.
\newblock In {\em CVPR}, 2018.

\bibitem{dai2017temporal}
X.~Dai, B.~Singh, G.~Zhang, L.~S. Davis, and Y.~Q. Chen.
\newblock Temporal context network for activity localization in videos.
\newblock In {\em ICCV}, 2017.

\bibitem{de2016online}
R.~De~Geest, E.~Gavves, A.~Ghodrati, Z.~Li, C.~Snoek, and T.~Tuytelaars.
\newblock Online action detection.
\newblock In {\em ECCV}, 2016.

\bibitem{deng2009imagenet}
J.~Deng, W.~Dong, R.~Socher, L.-J. Li, K.~Li, and L.~Fei-Fei.
\newblock {ImageNet}: A large-scale hierarchical image database.
\newblock In {\em CVPR}, 2009.

\bibitem{caba2015activitynet}
B.~G. Fabian Caba~Heilbron, Victor~Escorcia and J.~C. Niebles.
\newblock Activitynet: A large-scale video benchmark for human activity
  understanding.
\newblock In {\em CVPR}, 2015.

\bibitem{gao2017red}
J.~Gao, Z.~Yang, and R.~Nevatia.
\newblock {RED}: Reinforced encoder-decoder networks for action anticipation.
\newblock In {\em BMVC}, 2017.

\bibitem{gao2017turn}
J.~Gao, Z.~Yang, C.~Sun, K.~Chen, and R.~Nevatia.
\newblock {TURN TAP}: Temporal unit regression network for temporal action
  proposals.
\newblock {\em ICCV}, 2017.

\bibitem{gao2018dynamic}
M.~Gao, R.~Yu, A.~Li, V.~I. Morariu, and L.~S. Davis.
\newblock Dynamic zoom-in network for fast object detection in large images.
\newblock In {\em CVPR}, 2018.

\bibitem{girshick2015fast}
R.~Girshick.
\newblock {Fast R-CNN}.
\newblock In {\em ICCV}, 2015.

\bibitem{girshick2014rich}
R.~Girshick, J.~Donahue, T.~Darrell, and J.~Malik.
\newblock Rich feature hierarchies for accurate object detection and semantic
  segmentation.
\newblock In {\em CVPR}, 2014.

\bibitem{he2016deep}
K.~He, X.~Zhang, S.~Ren, and J.~Sun.
\newblock Deep residual learning for image recognition.
\newblock In {\em CVPR}, 2016.

\bibitem{hoai2014max}
M.~Hoai and F.~De~la Torre.
\newblock Max-margin early event detectors.
\newblock In {\em IJCV}, 2014.

\bibitem{hochreiter1997long}
S.~Hochreiter and J.~Schmidhuber.
\newblock Long short-term memory.
\newblock {\em Neural Computation}, 1997.

\bibitem{ioffe2015batch}
S.~Ioffe and C.~Szegedy.
\newblock Batch normalization: Accelerating deep network training by reducing
  internal covariate shift.
\newblock {\em arXiv:1502.03167}, 2015.

\bibitem{THUMOS14}
Y.-G. Jiang, J.~Liu, A.~Roshan~Zamir, G.~Toderici, I.~Laptev, M.~Shah, and
  R.~Sukthankar.
\newblock {THUMOS} challenge: Action recognition with a large number of
  classes.
\newblock \url{http://crcv.ucf.edu/THUMOS14/}, 2014.

\bibitem{karpathy2014large}
A.~Karpathy, G.~Toderici, S.~Shetty, T.~Leung, R.~Sukthankar, and L.~Fei-Fei.
\newblock Large-scale video classification with convolutional neural networks.
\newblock In {\em CVPR}, 2014.

\bibitem{kingma2014adam}
D.~P. Kingma and J.~Ba.
\newblock Adam: A method for stochastic optimization.
\newblock {\em arXiv:1412.6980}, 2014.

\bibitem{ma2016learning}
S.~Ma, L.~Sigal, and S.~Sclaroff.
\newblock Learning activity progression in lstms for activity detection and
  early detection.
\newblock In {\em CVPR}, 2016.

\bibitem{mnih2014recurrent}
V.~Mnih, N.~Heess, A.~Graves, et~al.
\newblock Recurrent models of visual attention.
\newblock In {\em NIPS}, 2014.

\bibitem{mnih2015human}
V.~Mnih, K.~Kavukcuoglu, D.~Silver, A.~A. Rusu, J.~Veness, M.~G. Bellemare,
  A.~Graves, M.~Riedmiller, A.~K. Fidjeland, G.~Ostrovski, et~al.
\newblock Human-level control through deep reinforcement learning.
\newblock 2015.

\bibitem{ren2015faster}
S.~Ren, K.~He, R.~Girshick, and J.~Sun.
\newblock {Faster R-CNN}: Towards real-time object detection with region
  proposal networks.
\newblock In {\em NIPS}, 2015.

\bibitem{shou2017cdc}
Z.~Shou, J.~Chan, A.~Zareian, K.~Miyazawa, and S.-F. Chang.
\newblock {CDC}: Convolutional-de-convolutional networks for precise temporal
  action localization in untrimmed videos.
\newblock In {\em CVPR}, 2017.

\bibitem{shou2018online}
Z.~Shou, J.~Pan, J.~Chan, K.~Miyazawa, H.~Mansour, A.~Vetro, X.~Giro-i Nieto,
  and S.-F. Chang.
\newblock Online action detection in untrimmed, streaming videos-modeling and
  evaluation.
\newblock In {\em ECCV}, 2018.

\bibitem{shou2016temporal}
Z.~Shou, D.~Wang, and S.-F. Chang.
\newblock Temporal action localization in untrimmed videos via multi-stage
  cnns.
\newblock In {\em CVPR}, 2016.

\bibitem{simonyan2014very}
K.~Simonyan and A.~Zisserman.
\newblock Very deep convolutional networks for large-scale image recognition.
\newblock {\em arXiv:1409.1556}, 2014.

\bibitem{sutton2018reinforcement}
R.~S. Sutton and A.~G. Barto.
\newblock {\em Reinforcement learning: An introduction}.
\newblock MIT press, 2018.

\bibitem{tran2015learning}
D.~Tran, L.~Bourdev, R.~Fergus, L.~Torresani, and M.~Paluri.
\newblock Learning spatiotemporal features with 3d convolutional networks.
\newblock In {\em ICCV}, 2015.

\bibitem{wang2016temporal}
L.~Wang, Y.~Xiong, Z.~Wang, Y.~Qiao, D.~Lin, X.~Tang, and L.~Van~Gool.
\newblock Temporal segment networks: Towards good practices for deep action
  recognition.
\newblock In {\em ECCV}, 2016.

\bibitem{wu2018blockdrop}
Z.~Wu, T.~Nagarajan, A.~Kumar, S.~Rennie, L.~S. Davis, K.~Grauman, and
  R.~Feris.
\newblock Blockdrop: Dynamic inference paths in residual networks.
\newblock In {\em CVPR}, 2018.

\bibitem{wu2018adaframe}
Z.~Wu, C.~Xiong, C.-Y. Ma, R.~Socher, and L.~S. Davis.
\newblock Adaframe: Adaptive frame selection for fast video recognition.
\newblock {\em arXiv:1811.12432}, 2018.

\bibitem{xu2017r}
H.~Xu, A.~Das, and K.~Saenko.
\newblock {R-C3D}: Region convolutional 3d network for temporal activity
  detection.
\newblock In {\em ICCV}, 2017.

\bibitem{xu2018temporal}
M.~Xu, M.~Gao, Y.-T. Chen, L.~S. Davis, and D.~J. Crandall.
\newblock Temporal recurrent networks for online action detection.
\newblock {\em arXiv:1811.07391}, 2018.

\bibitem{zhao2017temporal}
Y.~Zhao, Y.~Xiong, L.~Wang, Z.~Wu, X.~Tang, and D.~Lin.
\newblock Temporal action detection with structured segment networks.
\newblock In {\em ICCV}, 2017.

\end{thebibliography}
